\documentclass{article}

\usepackage[final]{nips_2017} % produce camera-ready copy

\usepackage{nips_2017}

\usepackage[utf8]{inputenc} % allow utf-8 input
\usepackage[T1]{fontenc}    % use 8-bit T1 fonts
\usepackage{hyperref}       % hyperlinks
\usepackage{url}            % simple URL typesetting
\usepackage{booktabs}       % professional-quality tables
\usepackage{amsfonts}       % blackboard math symbols
\usepackage{nicefrac}       % compact symbols for 1/2, etc.
\usepackage{microtype}      % microtypography
\usepackage{graphicx}       % graphics
\usepackage{float}          % figures
\usepackage{enumitem}       % lettered enumerate

\title{Sample-Efficient Reinforcement Learning through Transfer and Architectural Priors}

\author{
  Benjamin Spector\thanks{Presently a student at Horace Mann School.} \\
  SE(3) Group\\
  Cornell Tech\\
  New York, NY 10044\\
  \texttt{benjaminfspector@gmail.com}\\
  \And
  Serge Belongie\\
  SE(3) Group\\
  Cornell Tech\\
  New York, NY 10044\\
  \texttt{sjb344@cornell.edu}\\
}

\begin{document}

\maketitle

\begin{abstract}
  Recent work in deep reinforcement learning has allowed algorithms to learn complex tasks such as Atari 2600 games just from the reward provided by the game, but these algorithms presently require millions of training steps in order to learn, making them approximately five orders of magnitude slower than humans. One reason for this is that humans build robust shared representations that are applicable to collections of problems, making it much easier to assimilate new variants. This paper first introduces the idea of automatically-generated game sets to aid in transfer learning research, and then demonstrates the utility of shared representations by showing that models can substantially benefit from the incorporation of relevant architectural priors. This technique affords a remarkable 50x positive transfer on a toy problem-set.
\end{abstract}

\section{Introduction}

Consider the Atari 2600 game Seaquest \cite{seaquest}, one of the problems for which reinforcement learning is very effective, and visualized in figure 1. \cite{mnih2013}

Imagine a person learning to play this game. While individuals may apply different strategies, these are steps most decent players would take:

\begin{itemize}
    \item They would immediately recognize the existence of several sea creatures as well as a submarine, and likely ignore the rest of the water.
    \item They would jiggle the controls and see what happens, and quickly determine that they could only directly affect the yellow submarine. Additionally, this could probably be inferred from the fact that submarines differ from all other objects on-screen in that they are man-made.
    \item They would perceive the existence of a bar labeled ``oxygen'' and would deduce that, because the submarine they control is underwater and the bar slowly turns red with time, oxygen is a consumable resource that the player will need to replenish, and furthermore would know oxygen can be obtained from the atmosphere above the water.
\end{itemize}

\begin{figure}%[H]
    \centering
    \includegraphics[width=0.5\linewidth]{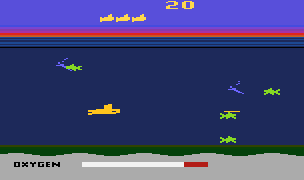}
    \caption{A snapshot from Seaquest, in which the player controls a yellow submarine and must dodge and destroy enemies as well as ensure that oxygen doesn't run out.}
\end{figure}

This list of relevant human intuitions and priors, even for a game as relatively simple as Seaquest, could probably be extended for several additional pages. In fact, in comparison to the raw pixel values that a reinforcement learning algorithm would be given, it almost seems that humans solve a fundamentally different problem than does machine reinforcement learning. Reinforcement learning must simply figure out how to associate large images with outputted actions to maximize a given scalar reward. Humans appear to already know the structure of both the problem and the solution before they even start playing, and just need to match the two together, which is a far easier task.

And, though the problems are in fact the same (for the computer and the human are in reality given identical observations and rewards), considering machine and human variants of reinforcement learning as solving different problems yields useful perspective: humans may be better than computers at general reinforcement learning, but the more significant effect is that they are very good at seeing a given problem as one they have already solved, bypassing the issue of learning.

Broadly, this means that humans make better assumptions about the problems that we care about than do reinforcement learning algorithms, but the key is that these assumptions are not innate but learned. So, in order to build computers that learn as quickly as people do, we ought to take a lesson from human players, and that is the goal of this project: to illustrate how building shared feature representations can drastically reduce training times when a new but related problem is encountered. \cite{pan2010}

This paper shows that in the context of a simple set of reinforcement learning problems with known relationships, an appropriate representation enables 50x positive transfer between variants. Because the problems used are simple, this work is intended primarily to illustrate the usefulness of the techniques outlined, with the potential for enormous training speedups if properly applied to more complex problems.

\section{Background}

This section summarizes the key ideas on which this paper builds. Section 2.1 provides context for reinforcement learning as used in this paper, focusing specifically on Q-learning and transfer learning. Section 2.2 introduces value iteration networks, the key architecture used in this paper.

\subsection{Reinforcement Learning}

Broadly, reinforcement learning is the act of acquiring skill at a task given only positive and negative reinforcement for previous actions in the form of reward. A reinforcement learning task is usually a Markov decision process (MDP), which is often termed an environment. Learning is usually subdivided into episodes, where the end of an episode marks a discontinuity in the underlying state of the MDP. An episode consists of a sequence of steps, during which the player is provided with an observed state and asked to return an action, upon which it is given its reward and the following observed state.

\subsubsection{Q-Learning}

Q-learning is a common way to solve the MDPs that encompass most games. The essential idea of Q-learning is that if, given a state, we can correctly assign some value to each action (assuming a finite action space), we can simply choose the highest-valued action in order to solve the MDP. More concretely, we define this using the Bellman equation $Q(s,a) = r + \gamma \; max \; Q(s', a')$, which states that the Q-value for a state is equal to the expected reward plus some discount factor times the maximum Q value in the following state, which accounts for the fact that immediate rewards are preferred because they have less uncertainty. Q-learning in its tabular form is known to converge to a global optimum strategy, but is ineffective for the set of environments used in this paper because each individual starting position is played only once, so that an arbitrary state is unlikely to ever be observed more than once or twice. \cite{watkins1992}

Deep Q Networks (DQNs) are a related extension of this idea. Because tabular Q-learning is intractable for large state spaces, DQNs introduce a neural network as a Q-function interpolator, allowing the algorithm to generalize to unseen states. Additionally, DQNs usually bring training offline, meaning that they store experiences and learn on them in batches rather than learning from observations and rewards directly as they are received. This has been found to increase sample efficiency and training stability compared to other methods, but is still profoundly inefficient in comparison to humans, requiring millions of frames to learn what a human may learn in hundreds. \cite{mnih2013}

Double DQNs (DDQNs) represent an improvement on DQNS by reducing the Q-value overestimation found in traditional DQNs. DDQNs solve this problem by swapping the action-selection and value-estimation networks repeatedly while the network is being updated, preventing overestimation loops from occurring due to the $max$ function in the Bellman equation. \cite{van2016}

\subsubsection{Transfer Learning}

Transfer learning is the use of knowledge gained in learning a source task to improve the learning of a different target task. There has been considerable interest in transfer learning because it is widely seen as a step necessary to producing artificial intelligence that behaves more similarly to humans. Humans are stateful entities, who are affected by what they learn (often needing just one example to change their approach to a new task), and it follows that algorithms which behave like humans should share this characteristic too. \cite{brown1988}

Transfer learning may improve the characteristics of learning the target task in three ways: it may increase the starting reward on the target domain, the rate of learning (slope) for the target task, or the maximum reward achievable. This work focuses on the second method. \cite{torrey2009}

Within online learning, one method of transfer learning on arbitrary MDPs is the use of Proto-value functions (PVFs), which reuse useful eigenvectors from the source domain to modify a target. Importantly, these are reward-independent, meaning that they encode knowledge about the task itself rather than just how to maximize its reward. \cite{ferguson2006}

Initial progress in transfer learning with DQNs copied the first layers from a known task to an unknown task, the idea being to have low-level features transfer between tasks. It was found that positive transfer was often observed, but not universally. A limitation of this approach comes from the model-free nature of DQNs, meaning that they are likely to latch on to any useful correlation regardless of whether or not it would be generally applicable. So, any positive transfer comes from the DQN happening upon a feature that does transfer well to a new task, but there are no guarantees provided about this occurring, and it was additionally found that even tasks that were expected to have positive transfer, Demon Attack and Space Invaders, due to their satisfying the similarity requirements outlined, had negative transfer instead. \cite{du2016}

Further progress in transfer learning used evolutionary algorithms combined with a novel modular neural network, called PathNet. It was found that an appreciable positive transfer of 1.33x was observed with optimal hyper-parameters on the Atari domain. As the problem domain used is considerably more complex than the one explored in this paper, this is a significant result. However, despite the positive transfer obtained, the number of training steps required to learn the game was not reduced due to the generally lower sample efficiency of PathNet in comparison with traditional reinforcement learning techniques, meaning that human-level learning rates are still far away for this class of MDPs. \cite{fernando2017}

\subsection{Value Iteration Networks}

Value iteration networks (VINs) provide a mechanism for neural networks to learn to plan by encapsulating an iterated version of the aforementioned Bellman equation inside an iterated process of convolution and max-pooling along the filter layer. The networks are limited to specific types of state-spaces, but have the desirable property that only the reward and attention models must be filled in in order to construct the network, making VINs an excellent candidate for a low-dimensional shared representation. \cite{tamar2016}

\section{Methods}

In this work, a problem-set is first defined, in which the key contribution is to automatically generate MDPs varying by a known transformation, allowing for greater ease of controlled transfer learning research. This problem-set is then solved by a custom-designed DDQN with an embedded shared representation to demonstrate the usefulness of these architectural priors in robust transfer.

\subsection{Environment}

The environment used in this paper is a type of grid-world, played on an 8x8 grid. In its simplified version (pictured in figure 2), the state is an 8x8x5 array, with the player being fed one-hot vectors to represent the different types of sites found on the grid (in other words, the object-space of the game). These are (as visualized):
\begin{itemize}
    \item Black (``Empty'')
    \item Blue (``Player'')
    \item White (``Target''): +3 reward; ends episode.
    \item Green (``Attractor''): +1 reward; disappears after passing over.
    \item Red (``Repulsor''): -1 reward; disappears after passing over.
\end{itemize}

Each step, the player selects and submits an action $a \in \{UP, DOWN, LEFT, RIGHT\}$, is moved accordingly, and then receives their reward and finally the new state. An episode may also end if the player moves off of the map, yielding a reward of -3, or if one-hundred turns have passed, yielding zero reward.

\begin{figure}
    \centering
    \includegraphics[width=0.3\linewidth]{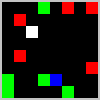}
    \caption{A snapshot of the simplified environment.}
\end{figure}

The existence of the off-map negative reward makes the grid-world slightly more difficult to learn because most of the signal to the network at the beginning of its training comes from it wandering off of the map, meaning that it first must learn to not do so before it can begin to target the other rewards.

\begin{figure}
    \centering
    \includegraphics[width=0.4\linewidth]{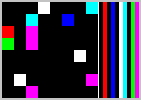}
    \caption{A snapshot of the auto-generated environment. Note the bars on the side of the image, which if read from left to right in the order, ``Player, Target, Attractor(s), Repulsor(s)'' describe the rules of that instance of the game. So, in this snapshot, the player is red, the target blue, the attractors white and cyan, and the repulsors green and magenta.}
\end{figure}

There is also a second version of this environment, the auto-generated variant (seen in figure 3). In this environment, the player receives 8 channels rather than 5, and there may be multiple types of +1 and -1 sites. Most importantly, the ordering of the channels is no longer guaranteed other than that black is always an empty site -- every automatically-generated version of the game will have some different assignment of colors to game object-types, and not all eight channels will necessarily be used in every game.

Rather than defining an individual MDP, the auto-generated environment describes a specific set of MDPs, and this makes it useful for investigating transfer learning because the differences between individual MDPs can be characterized precisely. One might see these sets of MDPs as a sort of labeled dataset for transfer learning algorithms. Furthermore, nearly limitless related MDPs can be efficiently generated on-demand, and because the simplified game is simply one such game with three channels removed, it can be seen as compatible with the auto-generated environment. These environments were integrated with the OpenAI Gym to allow its associated library keras-rl to be used. \cite{brockman2016}\cite{plappert2016}

\subsection{DDQN Architecture}

The architecture of the DDQN used in these experiments is outlined in figure 4, and is trained using a greedy-epsilon policy linearly annealed from 1 to 0.1 during training and held at 0.05 for testing.

\begin{figure}
    \centering
    \includegraphics[width=0.5\linewidth]{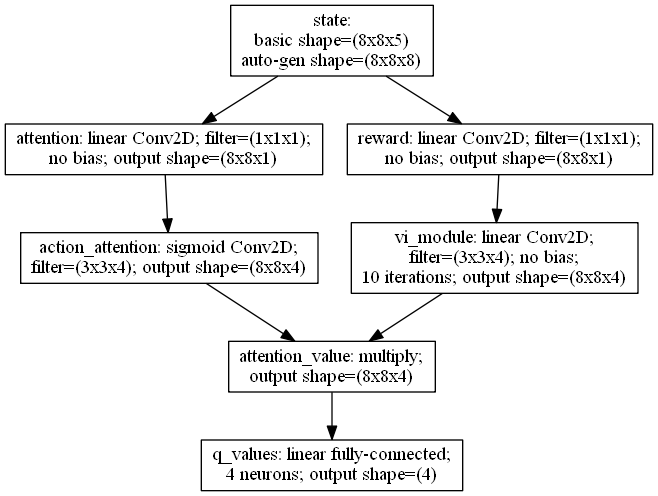}
    \caption{The architecture of the DDQN used for the following experiments. The auto-generated version has an six additional parameters compared to the simplified version due to its additional input channels, with three in each of the attention and reward layers.}
\end{figure}

These DDQNs are designed specifically for the environments provided, because they feature several important characteristics for the self-transfer and auto-generation transfer experiments to come.

First, both of the initial layers are 1x1x1 convolutions. This quickly brings the game-space into a prior-defined lower-dimensional space which represents just reward and attention. These two layers have 10 parameters together in the case of the simplified environment and 16 in the case of the auto-generated environment. This is similar to the technique described in \cite{pan2008}, but the lower-dimensional shared space is described through neural network architecture rather than semidefinite programs and principal component analysis, which is important because it preserves the geometry of the MDP within the representation, simplifying the process of learning the MDP.

Second, the model is actually quite deep despite its small size due to the unrolling of the value iteration model, and furthermore has variety within its layers, which allows greater insight into how the different parts of the model behave when perturbed.

Third, the entire model is quite small, which allows these experiments to be performed quickly and for little cost.

\section{Experiments and Results}

\subsection{Self-Transfer}

The first set of experiments involve the idea of self-transfer. The procedure is as follows: first, a DDQN is trained on the simplified environment end-to-end and its performance is measured. Then, some part of the network is re-initialized and the weights of all of the other layers are frozen. The network is then retrained and its performance measured over time. The purpose of this is to explore the robustness of the layer and its gradient, which in turn gives insight into candidate structures for networks with architecturally-defined shared representations. In other words, layers that are easily retrained are likely options for the problem-specific parts of a network, whereas layers that are much more difficult to retrain should be used as general parts of the network.

The original network was trained over the course of 100,000 steps, and it was found that some parts of the network are distinctly more difficult to retrain than others.

The easiest layers of the network to retrain are unsurprisingly the reward and attention models, in that order. The reward model can reach its original level of performance within 600 training steps, and the attention model requires approximately 1,000.

Somewhat more difficult to retrain was the action\_attention layer. This is probably because it has more parameters than either of the attention or reward models in a larger convolution, but it nonetheless quite quickly from the beginning and achieved an average testing reward of 2.5 after approximately 2,000 steps, and a slow improvement from there on.

The q\_values layer took significantly longer to retrain. This is likely because of its number of parameters, as it accounts for about 90\% of the model's total. However, because it only needs to produce a linear association between the attention\_value layer and the actual Q values, it achieves its original reward within approximately 10,000 steps.

The value iteration module was found to be the most difficult to re-train, which could not relearn weights yielding a reward even close to the network's original reward. This may be due to the iterated nature of the module causing the model to become brittle and highly sensitive to its exact weights and making gradient descent difficult.

Complete hyper-parameter settings and training histories can be found in Appendix A.

\subsection{Auto-generation transfer}

The second set of experiments work with the problem of transfer learning between different auto-generated games. Crucially, because the underlying rules of the game are identical, and only the appearance of the game has changed, the only layers that need to be relearned for any of the thousands of possible games are the reward and attention layers (all others being frozen), because they alone map the game-space into the common reward and attention spaces.

Two games were selected for this experiment, seeds 42 and 256, shown in figure 5.

\begin{figure}[H]
    \centering
    \includegraphics[width=0.5\linewidth]{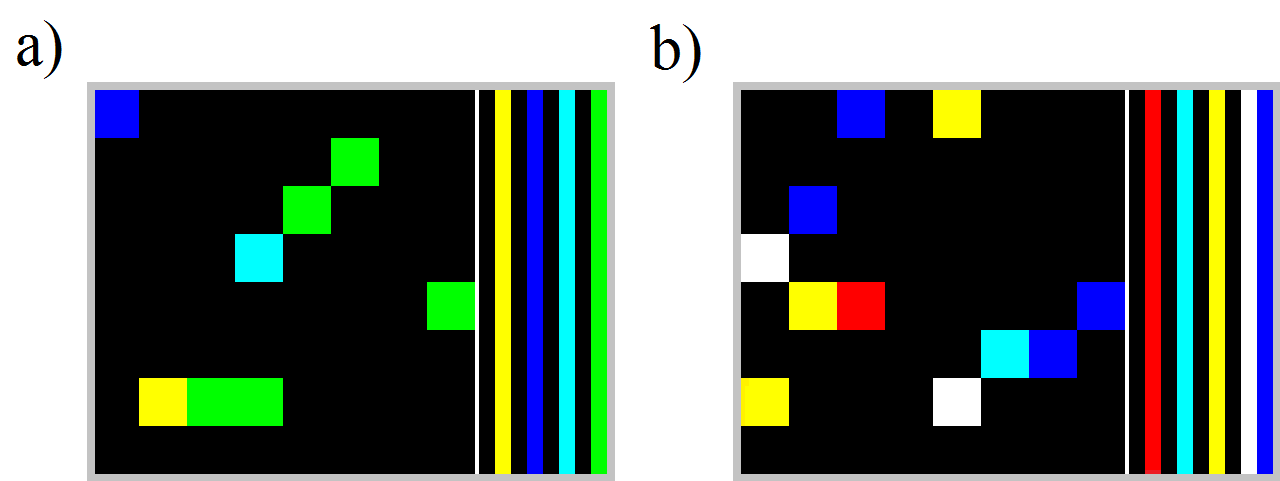}
    \caption{A visualization of the auto-generated games from seeds (a) 42 and (b) 256. Seed 42 is used for the control, and seed 256 is trained with transfer.}
\end{figure}

The reader should note that the results displayed were found to generalize to arbitrary pairs of games. Seeds 42 and 256 were chosen as a representative example simply because they were quite different from each other, and also because they would likely pose challenges for transfer learning agents. First, seed 256 is decidedly more difficult because it has two filters mapped to the repulsor object, whereas seed 42 has a one-to-one mapping for all object types. Second, between the seeds some filters are reused but change meaning, and others are entirely new. Because no filter retains its original meaning, the network must relearn each mapping in order to successfully play the new game -- it could not, for example, keep the attention block and relearn just the reward block.

Transfer between pairs of auto-generated games was found to be very effective. Comparison of figures 6 and 7a show that the rate of training new games is drastically increased compared to the rate of training the first game, even with the same learning rate. Specifically, the transferred model reached a reward of 2 approximately 50 times faster than the end-to-end (non-transferred) model, showing that robust shared representations are indeed effective for transfer learning between games. 

\begin{figure}
    \centering
    \includegraphics[width=0.5\linewidth]{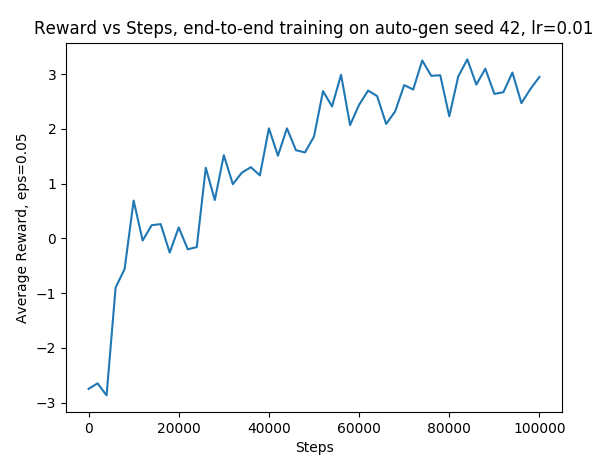}
    \caption{The training history of the control model on seed 42.}
\end{figure}

\begin{figure}[H]
    \centering
    \includegraphics[width=\linewidth]{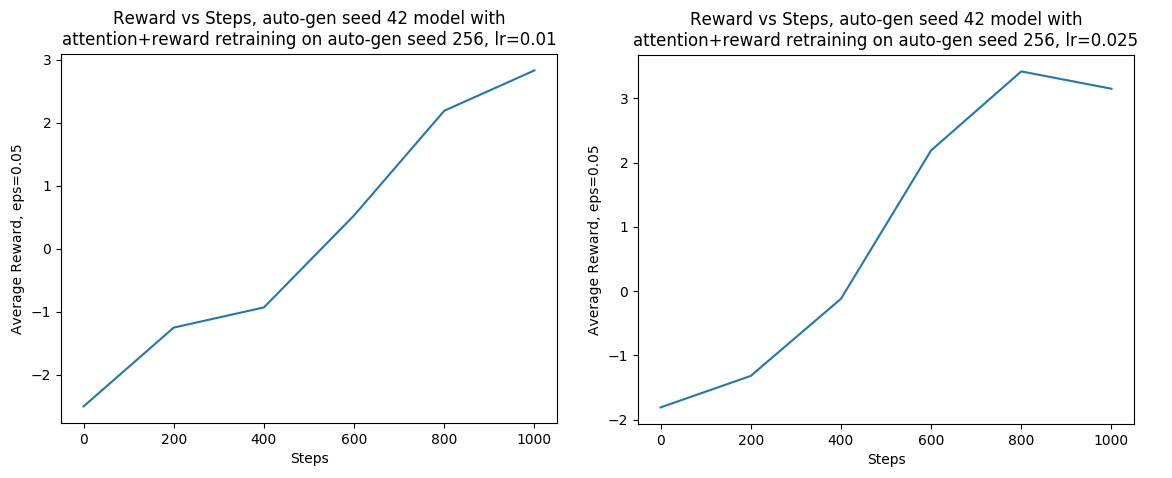}
    \caption{The training histories of the transferred model. Fewer data points are collected to minimize episode interruption in the model. Graph a) shows the training history at the same learning rate as used in the control; graph b) shows the training history at a higher learning rate.}
\end{figure}

\section{Discussion}

\subsection{Self-transfer}

Consider the shared representation chosen: beyond the attention and reward layers, the network never sees the original channels provided. All that the deeper layers ``know'' is how to solve the MDP given 8x8 grids of attention and reward. So, the key requirement of the shared representation is that there must exist some mapping between games and the representation itself such that when the representation is solved, the game is solved, too. Choosing a good representation therefore requires minimizing the complexity of the mappings from games to the representation, so that it is as general as possible, and the quality and depth of the shared representation itself, so that it is as effective as possible. A good representation concentrates all of the difficulty of training to general layers, so that task-specific layers can be learned quickly and easily.

So, a representation for which the general layers are much more difficult to train than the game-specific layers may be considered to be an effective representation, and a representation for which the general layers are much easier to train than the game-specific layers is proportionally less effective.

The self-transfer experiments reveal that the design of the network is effective for the set of MDPs the network between which is responsible for transferring because the layers that need to be relearned are the easiest to retrain by a significant margin, meaning that the network has offloaded most of the difficult-to-train computation to game-invariant layers.

\subsection{Auto-generation transfer}

The effectiveness of transfer learning may be measured by the amount of training steps required to learn a given MDP to a certain reward without the transfer in comparison with the time required to learn the same MDP to the same reward with the transfer. This is a good general metric, because it simply defines the ability of the transfer learning to reduce the complexity of new problems, and additionally is independent of the architecture of the transfer.

Under this metric, these results conclusively demonstrate that architectural priors may be highly effective, moving these auto-generated game variants from the class of traditional reinforcement-learning problems requiring thousands of training episodes to a near-human class requiring several dozen.

In a sense, the fact that re-learning sixteen linear parameters of a network is much faster than learning an 1100-parameter nonlinear network end-to-end is not only noncontroversial, but also obvious. It should be emphasized that this, is fact, the point -- transfer, if done properly, should make learning easy.

These results are distinct from those already seen in the field of transfer learning simply due to the degree of the positive transfer observed. In transfer for reinforcement learning, positive transfer results of several dozen percent are significant, so to the best of the knowledge of the author this work represents an order-of-magnitude greater positive transfer than ever previously demonstrated using neural networks.

Naturally, it is not yet known how to build such complete shared representations for more complex and diverse sets of problems, such as the benchmark Atari domain often used in prior works. However, the performance of humans on these tasks provides an upper bound for the minimum possible number of episodes required, showing that the creation of analogously robust priors is undoubtedly possible.

\section{Conclusions}

This paper introduced the idea of automatically-generated game sets to aid in transfer learning research, and then demonstrated the utility of architectural priors by showing that reinforcement learning agents can substantially benefit from the resulting capability for positive transfer, decreasing training times by 50x on an example problem-set.

Despite the simplicity of both the MDPs and the shared representation used, the underlying ideas extend far beyond. Returning to the example from the introduction, a person playing Seaquest would probably construct underlying attention and reward spaces as are used in this work. Naturally, several other layers to the representation would probably also be necessary, but this highlights an intermediate stage between model-free (DQN-like) and model-based algorithms, which might be called model-aware algorithms, for they recognize and make use of the structure of an underlying model through their architectural priors even though they are still intrinsically model-free. Additionally, this work further postulates that humans may also make use of model-aware type algorithms too.

There are three straightforward paths that could be taken to extend this work:
\begin{itemize}
    \item First, this work could be extended by further decreasing the number of steps required to play these grid world-type games into the realm of one-shot or few-shot learning. This might at first seem simple to achieve by increasing the learning rates and batch sizes for transfer; however, to go any further would likely require a more coherent and useful exploration criteria than greedy-epsilon, as suggested in \cite{taylor2009}.
    \item One could also extend this technique to more complex, higher-dimensional problems such as a restricted set of Atari games. For example, a representation consisting of semantic reward and attention maps as utilized in this paper might also be effective at transfer between Pong and Breakout due to the common mechanic of intercepting the path of a ball with a paddle. One could also dynamically generate these more complex tasks in analogue to the auto-generated game used in this paper.
    \item Last, one could undertake the challenge of learning the shared representation rather than encoding it in the network architecture. Based on the results presented in this paper, this may be a promising path to human-level learning rates on Atari-class games, but it is not yet known how one would accomplish this.
\end{itemize}

In conclusion, there is great potential to improve even state-of-the-art reinforcement learning techniques using these principles. It is therefore hoped that the methods and results presented in this work will encourage future work on transfer learning using neural networks with shared representations.

\section*{Acknowledgements}

The authors would like to thank the rest of SE(3) for useful conversations during the course of this research, and especially Zekun Hao and Baoguang Shi. We thank Joshua Gruenstein from the Small-Data Lab at Cornell Tech for countless reviews and sanity-checks, and Sat Chatterjee from Two Sigma Investments for his invariably useful suggestions.

%\nocite{*}
%\bibliographystyle{ieeetr}
%{\small \bibliography{ref}}

\newpage
\appendix

\begin{center}
\section{Self-transfer Complete Graphs}
\end{center}

The model trained end-to-end in figure 8 is used as the basis in the following experiments in this appendix, and follows nearly linear progress during training, in contrast to several of the following graphs.

\begin{figure}[H]
    \centering
    \includegraphics[width=0.5\linewidth]{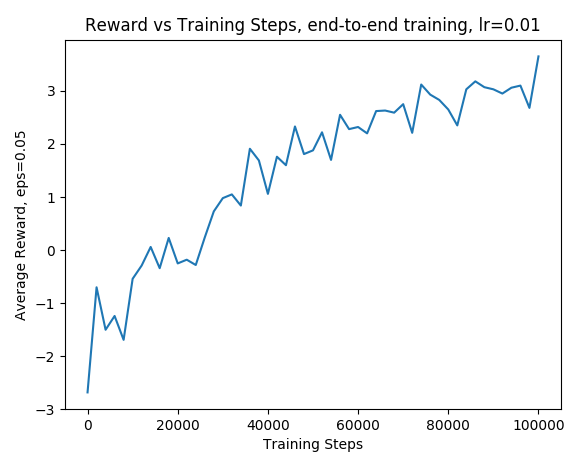}
    \caption{The end-to-end training history of the original model trained on the simplified game.}
\end{figure}

\begin{figure}
    \centering
    \includegraphics[width=.87\linewidth]{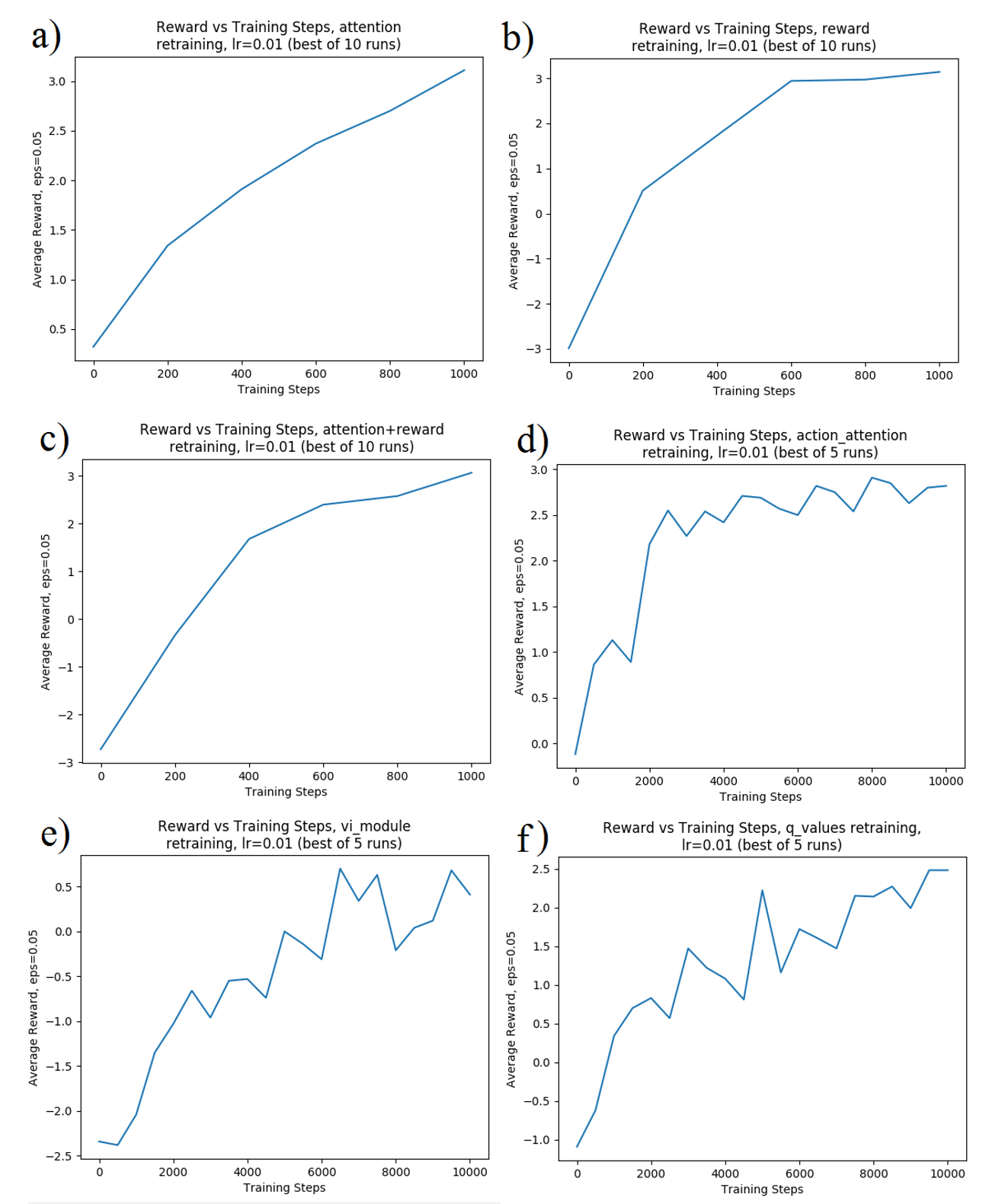}
    \caption{The full data behind the results of the self-transfer experiment, including hyper-parameter settings.}
\end{figure}

Figure 9 shows complete results from the self-transfer experiment, in which specific layers of the DDQN were retrained and the network's average reward achieved graphed over time. The results and implications of each graph are discussed below:

\begin{enumerate}[label=\alph*]
    \item The attention retraining process is quite straightforward and linear. As it is just eight linear parameters, it reaches its original performance very quickly. Intriguingly, the model actually appears to reach an average reward far above random actions with a random initialization of this layer, indicating that the layer may be less important to the network's performance than the reward layer.
    \item The reward retraining process is the quickest of all tested, increasing its average reward by over 3.5 during merely the first 200 steps, and converging at 600. This result is somewhat unexpected given that there are great non-linearities introduced further along in the network by the value iteration model, making it surprising that this layer learns so rapidly.
    \item The retraining of both the attention and reward layers is analogous to the auto-generated game transfer experiment in terms of layers retrained. It appears to train approximately as quickly as in the second transfer experiment, which reaffirms that the remainder of the network really is general to all possible auto-generated games.
    \item The action attention model is somewhat more difficult than any of the other models to retrain, but still converges very quickly. The slight increase in reward towards the end of its training can probably be accounted for by the decreasing epsilon parameter during training.
    \item The value iteration module was by any measure the most difficult layer to retrain. Not only did it consistently fail to reach its original level of performance, but its results also could not be improved by decreasing the learning rate or training it for extended periods of time -- one experiment even ran it to 250,000 steps, or significantly longer than the time for which original network was trained.
    \item The final q\_values layer was found to be fairly straightforward and reliable to train. As it accounts for over 1,000 of the parameters of the network's approximately 1,100 total parameters, it is unsurprising that it takes considerably longer to re-train than any of the early layers of the network (with the notable exception of the value iteration module). Due to the attention model, this can be seen as an issue of exploration, because a given weight parameter can only be trained when the network has a non-zero value in that spot, which occurs only near the actual location of the player tile. Given the geometric symmetries of the layer, there are probably other training techniques that could rapidly discover good weights, such as those detailed in \cite{schmidhuber1997}, but when viewed purely from a back-propagation standpoint, there is probably little to be done to increase the learning rate of this layer due to the aforementioned problems.
\end{enumerate}

\end{document}